\documentclass{article} 
\usepackage{iclr2027_conference,times}

\usepackage{amsmath,amsfonts,bm}

\def\eqref#1{equation~\ref{#1}}

\def\1{\bm{1}}

\DeclareMathAlphabet{\mathsfit}{\encodingdefault}{\sfdefault}{m}{sl}
\SetMathAlphabet{\mathsfit}{bold}{\encodingdefault}{\sfdefault}{bx}{n}

\usepackage{hyperref}
\hypersetup{
    colorlinks=true,
    linkcolor=blue,
    citecolor=blue,
    urlcolor=magenta
}
\usepackage{url}
\usepackage{algorithm}
\usepackage{algorithmic}
\usepackage[table]{xcolor}

\usepackage{newfloat}
\usepackage{listings}

\usepackage{booktabs}
\usepackage{graphicx}
\usepackage{multirow}
\usepackage{amsmath}
\usepackage{amssymb}
\usepackage{wrapfig}

\title{LiG-DETR: Local-in-Global Reassembly in Latent Space for Aerial Object Detection}

\author{
Yupeng Zhang\textsuperscript{1}\quad
Fangzhuo Gao\textsuperscript{1}\quad  
Juntao Cheng\textsuperscript{3,4}\quad  
Ziyi Zhao\textsuperscript{2}\quad 
Liang Wan\textsuperscript{1}\quad
Ruize Han\textsuperscript{2}
\\
\textsuperscript{1}Tianjin University.
\qquad
\textsuperscript{2}Shenzhen University of Advanced Technology.
\qquad
\textsuperscript{3}Shanghai Jiaotong University.
\\
\textsuperscript{4}Shanghai AI Laboratory.
\\
{\small
\texttt{\{zhangyupeng, fangzhuo, lwan\}@tju.edu.cn, suat25060113@stu.suat-sz.edu.cn}
}
\\
{\small
\texttt{hanruize@suat-sz.edu.cn, juntao-cheng@sjtu.edu.cn}}
}

\iclrfinalcopy 

\begin{document}

\maketitle

\begin{abstract}
Aerial object detection faces substantial scale and density variations. Small objects are easily degraded by downsampling and feature compression, while medium and large objects require sufficient global context. Existing methods mainly follow two paradigms: image slicing provides clearer local evidence but relies on independent crop-level prediction and post-processing, whereas feature- and query-level optimization preserves unified inference but operates on already compressed full-image representations, limiting recovery of fine-grained information.
This raises a key question: can aerial detection directly acquire high-fidelity local evidence before feature degradation and integrate it into a unified end-to-end framework?
To this end, we propose LiG-DETR, an Efficient Global-Local Reassembly framework that reformulates image slicing as high-fidelity local feature acquisition. A shared encoder extracts global and locally magnified features, which are projected into the detector feature space. The projected local features are reassembled according to their original spatial locations to form a globally aligned local feature level, and a single DETR decoder jointly decodes global and local features. To reduce redundant computation, Context-Preserved Selective Reassembly focuses high-resolution encoding on informative regions while preserving a dense feature layout, and Density-Aware Adaptive Query Allocation adapts the decoder query budget using encoder proposal scores.
Experiments show substantial gains on small and medium objects while retaining strong large-object performance, with favorable accuracy--efficiency trade-offs and improved cross-domain generalization.
\textit{The code will be released.}
\end{abstract}



\section{Introduction}

Aerial object detection~\citep{du2018unmanned,du2019visdrone} is important for UAV perception and wide-area surveillance. Compared with ground-level imagery, aerial imagery typically covers larger fields of view and exhibits greater variations in object scale and density, often containing many tiny objects alongside larger ones. Effective detection therefore requires fine-grained local evidence for small objects while preserving global context across scales. However, small objects are particularly vulnerable to degradation because their limited spatial extent provides weak texture, boundary, and shape cues that can be further suppressed by background clutter and feature compression.

Existing methods mainly address aerial small-object detection in two ways. Image slicing methods, such as SAHI~\citep{akyon2022slicing} and ASAHI~\citep{zhang2023adaptive}, improve small-object visibility by cropping and resizing local regions. Other methods better exploit full-image representations via feature or query manipulation, including multi-scale fusion with FPN~\citep{lin2017feature}, sparse high-resolution computation with QueryDet~\citep{yang2022querydet}, attention enhancement with Deformable DETR~\citep{zhu2020deformable} and TPH-YOLOv5~\citep{zhu2021tph}, and density- or query-aware refinement with DQ-DETR~\citep{huang2024dq} and Dome-DETR~\citep{hu2025dome}.
These two paradigms reveal a design trade-off. Image slicing acquires clearer local visual evidence, but requires independent crop-level detection and subsequent box fusion or NMS-like post-processing, leading to redundant computation, duplicate predictions, and fragmented inference. In contrast, feature- or query-level refinement preserves unified end-to-end inference, but operates on already encoded full-image representations. Once fine-grained evidence is weakened by downsampling and feature compression, subsequent representation enhancement may be unable to fully recover it.

\begin{wrapfigure}{r}{0.48\columnwidth}\vspace{-0pt}
    \centering
    \includegraphics[width=\linewidth]{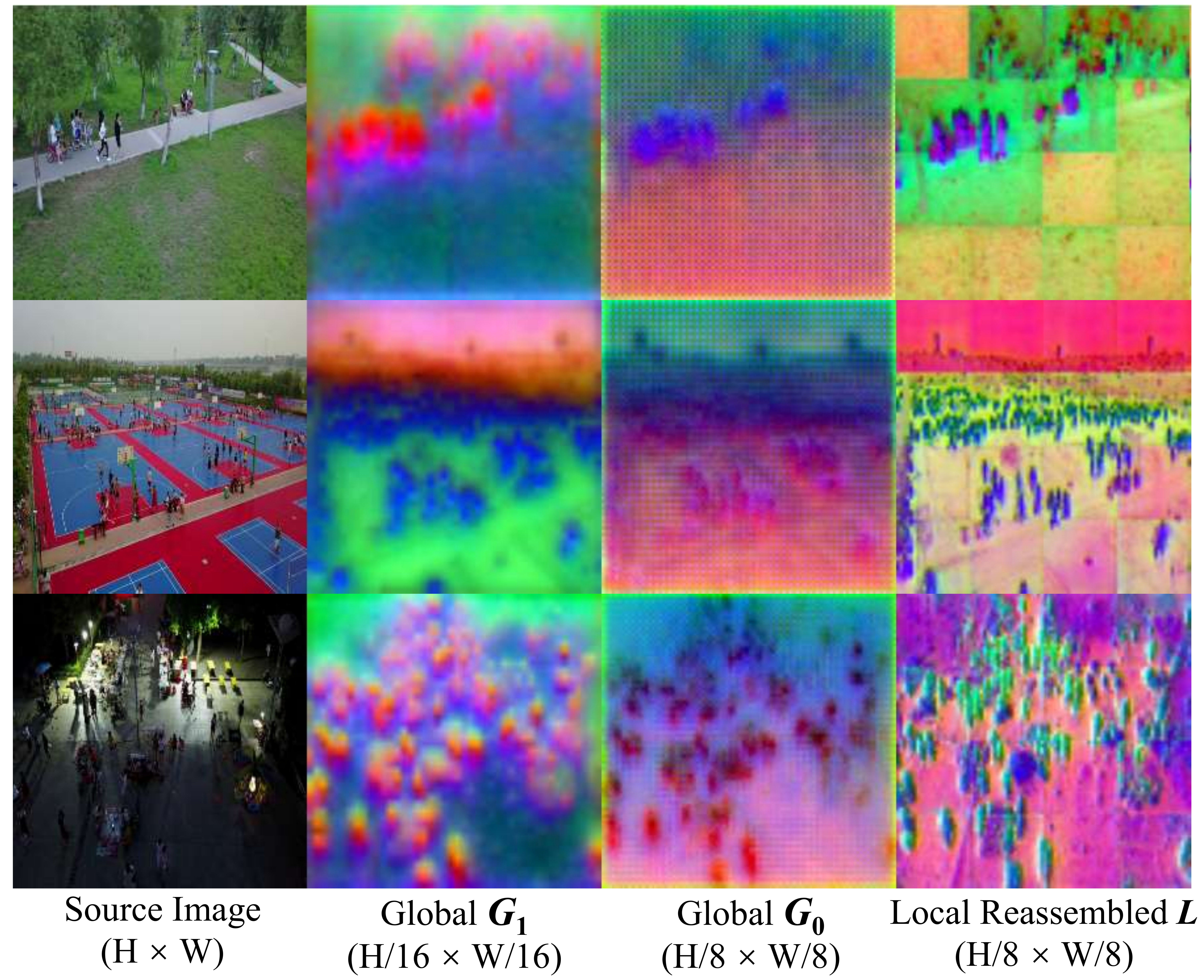}\vspace{-22pt}
    \caption{Comparison of small-object evidence under global and local encoding. Full-image features (Global $G_1$/$G_0$) exhibit sparse or ambiguous small-object responses. Considering locally magnified encoding preserves clearer evidence before feature degradation, we propose reassembled local feature $L$ retaining high-resolution evidence in global coordinate space.}  
    \vspace{-8pt}
    \label{fig:distribute}
\end{wrapfigure}

To further examine the limitations of full-image representations for small-object detection, we analyze representative DETR-based detectors~\citep{zhu2020deformable,zhao2024detrs,robinson2026rf}. Despite their strong global context modeling and end-to-end prediction capabilities, full-image encoding can weaken the fine-grained information of small objects. Repeated downsampling and feature compression may leave only sparse and weak responses before decoding, limiting the visual evidence available to subsequent feature- or query-level refinement.
We use RF-DETR~\citep{robinson2026rf} for diagnostic analysis and visualize its intermediate representations in Fig.~\ref{fig:distribute}. We denote the encoder-native global feature as Global $G_1$, and the \textit{higher-resolution global feature} generated through projection as Global $G_0$. As shown in Fig.~\ref{fig:distribute}, small-object responses in Global $G_1$ are sparse and coarse. Although Global $G_0$ provides higher spatial resolution, it is still derived from the full-image representation and may not fully recover fine-grained object information already weakened during encoding. 
This way, we consider to construct the Local Reassembled feature $L$, where local image regions are magnified and encoded by the same visual encoder, then projected and reassembled according to their original spatial locations (in the global coordinate space).
We can see that the locally magnified and reassembled feature $L$ preserves clearer and more complete evidence before such degradation occurs.
This observation raises a natural question: \textit{beyond refining existing representations}, can aerial object detection directly \textit{acquire high-fidelity local evidence before feature degradation and integrate it into a unified end-to-end detection framework} for stronger small-object representations?

Based on this observation, we propose \textbf{LiG-DETR}, an efficient \textbf{Local-in-Global} latent reassembly framework based on DETR that reformulates image slicing from an independent detection strategy into a mechanism for acquiring high-fidelity local evidence. Instead of predicting on each crop independently, LiG-DETR encodes locally magnified observations and reassembles their features into a unified global coordinate space. In this way, image slices become feature acquisition units rather than prediction units. The reassembled local evidence and global features are jointly decoded by a single DETR decoder, eliminating crop-level detection heads, box fusion, and additional NMS.

Specifically, LiG-DETR uses a shared encoder for the global branch and the local high-resolution branch. The \textbf{Global-Local Reassembly Projector} maps global and local features into the detector feature space, after which the projected local features are reassembled according to their original spatial locations to form a globally aligned local feature level. Within this framework, \textbf{Context-Preserved Selective Reassembly (CPSR)} concentrates high-resolution encoding on informative regions and fills unselected locations with corresponding global features to preserve a dense feature layout, whereas \textbf{Density-Aware Adaptive Query Allocation (DA-AQA)} adapts the decoder query budget to image content according to encoder proposal scores. Together, they reduce redundant computation along the spatial and query dimensions, respectively.
Experiments on VisDrone-DET2019, AI-TOD-v2, and DGDet-VisDrone (built in this work) demonstrate that LiG-DETR obtains substantial gains on small and medium objects while retaining strong large-object performance, with favorable accuracy-efficiency trade-offs and improved cross-domain generalization.



\section{Related Work}

\textbf{Image-level Slicing for Aerial Small Objects.}
Image slicing is widely used to improve small-object visibility in aerial images by cropping and resizing local regions before detection. SAHI~\citep{akyon2022slicing} performs detection on overlapping image slices and merges mapped predictions afterwards, while ASAHI~\citep{zhang2023adaptive} adaptively adjusts slice sizes to reduce redundant processing. This paradigm improves local object visibility by increasing the effective image scale, but couples high-resolution observation with independent crop-level prediction and box fusion or NMS-like post-processing. In contrast, LiG-DETR decouples high-resolution evidence acquisition from crop-level prediction: local crops are used only for feature acquisition, and their features are reassembled into the global coordinate space for unified end-to-end decoding.

\textbf{Multi-scale Feature and Query Modeling for Aerial Detection.}
Aerial object detection requires representations robust to large scale and density variations. Multi-scale methods such as FPN~\citep{lin2017feature}, PANet~\citep{liu2018path}, and CE-FPN~\citep{luo2022fpn} fuse spatial details and semantic information across feature levels. For aerial and small-object detection, TPH-YOLOv5~\citep{zhu2021tph}, QueryDet~\citep{yang2022querydet}, YOLO-MS~\citep{chen2025yolo}, FBRT-YOLO~\citep{xiao2025fbrt}, and HS-FPN~\citep{shi2025hs} further improve high-resolution or multi-scale features through attention, sparse computation, and adaptive modeling.
DETR-based detectors, including Deformable DETR~\citep{zhu2020deformable}, DN-DETR~\citep{li2022dn}, DINO~\citep{zhang2022dino}, and RT-DETR~\citep{zhao2024detrs}, improve feature interaction, training, and query initialization, while efficient variants such as D-FINE~\citep{peng2025d}, DEIMv2~\citep{huang2025real}, and RF-DETR~\citep{robinson2026rf} improve the accuracy--efficiency trade-off. For dense aerial scenes, DQ-DETR~\citep{huang2024dq} adapts query number and locations to object density, while Dome-DETR~\citep{hu2025dome} combines density-guided feature processing, sparse foreground computation, and adaptive query initialization.
Overall, these methods mainly enhance full-image representations through feature or query modeling. However, once fine-grained visual information is lost during early downsampling and feature compression, later refinement can hardly recover it. In contrast, LiG-DETR directly acquires high-fidelity local visual evidence before such degradation and reintegrates it with global representations for unified end-to-end detection.

\section{The Proposed Method}

\subsection{Overview}
\begin{figure*}[t!]
    \centering
    \includegraphics[width=1\linewidth]{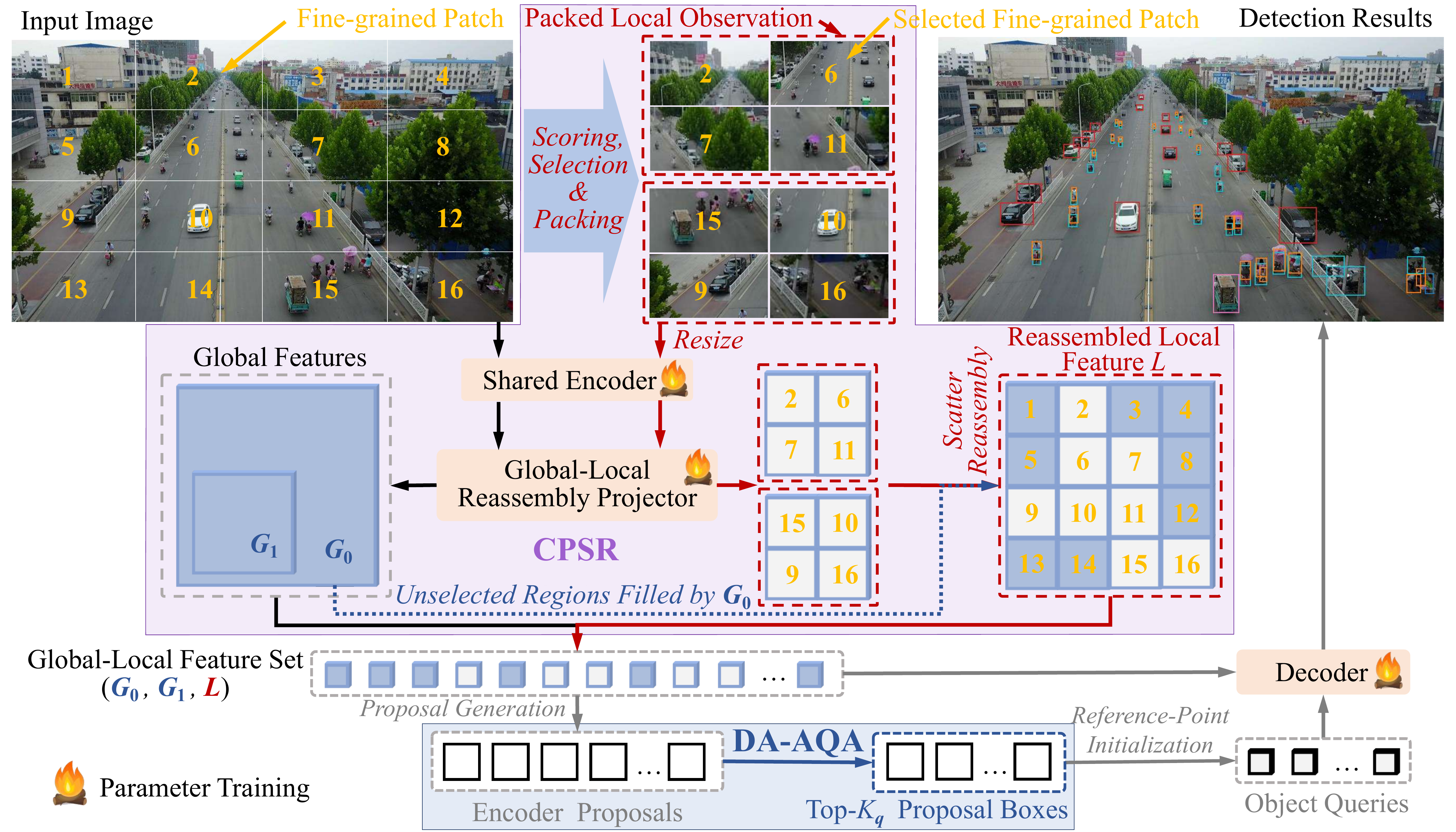}
    \vspace{-15pt}
    \caption{Overview of LiG-DETR. Black and red arrows denote the full-image global branch and the zoomed-in local branch, respectively. The full image is encoded into the encoder-native Global $G_1$, from which the higher-resolution Global $G_0$ is generated through projection. \textbf{CPSR} selects informative fine-grained patches for high-fidelity local evidence acquisition. The projected local features are spatially reassembled, while unselected regions are filled with the corresponding Global $G_0$ features, forming the context-preserved dense local feature $L$. The resulting feature set $\{G_0,G_1,L\}$ is used for proposal generation and unified decoding by a single DETR decoder. \textbf{DA-AQA} determines an image-specific query budget $K_q$ from encoder proposal scores and uses the top-$K_q$ proposal boxes to initialize decoder reference points.}
    \label{fig:overview}
    \vspace{-15pt}
\end{figure*}

Given an image $I\in\mathbb{R}^{3\times H\times W}$, LiG-DETR follows a DETR-style pipeline and predicts objects in the original image coordinate space. As shown in Fig.~\ref{fig:overview}, it consists of a global branch (black arrows), a zoomed-in local branch (red arrows), a Global-Local Reassembly Projector, and a single DETR decoder. The two branches share the same visual encoder, while decoding is performed only once for unified set prediction.
The global branch encodes the resized full image to preserve scene-level context, producing the encoder-native Global $G_1$, from which a higher-resolution Global $G_0$ is generated through projection. In parallel, the local branch acquires high-fidelity local evidence by encoding selected zoomed-in regions. The projected local features are reassembled according to their original spatial locations, while unselected regions are filled with the corresponding Global $G_0$ features, forming the context-preserved dense local feature $L$. The resulting feature set $\{G_0,G_1,L\}$ is then used for proposal generation and unified decoding by a single DETR decoder.
Within this framework, \textbf{Context-Preserved Selective Reassembly (CPSR)} concentrates high-resolution encoding on informative spatial regions, while \textbf{Density-Aware Adaptive Query Allocation (DA-AQA)} adapts the decoder query budget based on encoder proposal scores. Together, they reduce redundant computation across spatial regions and decoder queries.

\subsection{Global-Local Reassembly for High-Fidelity Local Evidence}

Full-image encoding is essential for global context modeling, but spatial compression can weaken the fine-grained visual cues of small objects. Rather than relying solely on full-image representations or independent crop-level detection, LiG-DETR acquires \textbf{high-fidelity local evidence} from magnified observations, reassembles it into the global spatial coordinate system, and jointly decodes it with global features using a single DETR decoder.

Let $\phi(\cdot)$ denote the shared visual encoder, and $\Pi_g(\cdot)$ and $\Pi_l(\cdot)$ the global and local projection operations in the Global-Local Reassembly Projector, respectively. For the global branch, we obtain {\footnotesize $\{G_0,G_1\}=\Pi_g(\phi(I))$}, where $G_1$ denotes the encoder-native global feature at stride 16, and $G_0$ is the higher-resolution global feature generated through projection at stride 8. For an $H\times W$ input, their spatial sizes are approximately $\frac{H}{16}\times\frac{W}{16}$ and $\frac{H}{8}\times\frac{W}{8}$, respectively. Thus, $G_0$ has $2\times$ higher spatial resolution along each axis and approximately $4\times$ more spatial tokens than $G_1$. 
However, both derive from the same full-image encoding, where fine-grained small-object evidence may be weakened.

To acquire local evidence before such degradation, we define a full local reassembly scheme. The input image is divided into $2\times2$ non-overlapping \textbf{local crops} $\{I_i^l\}_{i=1}^{4}$. Each crop is resized to the encoder input size and encoded by the shared visual encoder as {\footnotesize $Z_i^l=\phi(I_i^l)$}, for $i=1,\ldots,4$. This magnifies local content before encoding, allowing small objects to occupy more visual tokens without changing the encoder architecture. The resulting local features are projected and spatially reassembled as {\footnotesize $L_{\mathrm{full}}=\mathcal{R}\!\left(\{\Pi_l(Z_i^l)\}_{i=1}^{4}\right)$}, where $\mathcal{R}(\cdot)$ denotes deterministic spatial reassembly.

The reassembled feature $L_{\mathrm{full}}$ has the same spatial resolution as $G_0$, but contains high-fidelity evidence acquired from locally magnified observations. Together with $\{G_0,G_1\}$, it forms a globally aligned feature set for unified decoding in the original image coordinate space.

Full reassembly, however, requires high-resolution encoding for all local regions. Directly enlarging the full image suffers from a similar inefficiency: doubling both image dimensions increases the number of visual tokens by approximately $4\times$, and the quadratic self-attention cost by approximately $16\times$. Since informative content is spatially non-uniform in aerial images, high-resolution computation need not be allocated uniformly, motivating our selective reassembly strategy.

\subsection{Context-Preserved Selective Reassembly (CPSR)}

To reduce the cost of full reassembly, \textbf{Context-Preserved Selective Reassembly (CPSR)} selectively allocates high-resolution encoding to informative regions while preserving the same dense and globally aligned feature layout required by subsequent decoding.

Specifically, the input image is divided into $N\times N$ \textbf{fine-grained patches} $\{p_j\}_{j=1}^{N^2}$. To identify informative regions with minimal additional cost, we use a Laplacian-based high-frequency score. Given the resized and ImageNet-normalized input tensor $X$, we first average its three channels as {\footnotesize $g=\frac{1}{3}\sum_{c=1}^{3}X_c$}, and then apply a $3\times3$ four-neighbor Laplacian kernel. The importance score of patch $p_j$ is defined as the mean absolute Laplacian response over valid image pixels:
\begin{equation}
\footnotesize
a_j
=
\frac{
\sum_{u\in p_j} m(u)\left|(\mathbf{L}*g)(u)\right|
}{
\max\left(1,\sum_{u\in p_j}m(u)\right)
},
\end{equation}
where $m(u)\in\{0,1\}$ masks padding and $\mathbf{L}$ denotes the four-neighbor Laplacian kernel.

Given a keep ratio $\gamma$, we retain
{\footnotesize $K_p=\operatorname{clip}\!\left(\left\lceil\gamma N^2\right\rceil,K_{\min},K_{\max}\right)$}
patches and select
{\footnotesize $\mathcal{S}=\operatorname{TopK}\!\left(\{a_j\}_{j=1}^{N^2},K_p\right)$}.
The selected patches are sorted by importance and packed into compact $G\times G$ local observations, yielding {\footnotesize $B=\lceil K_p/G^2\rceil$} observations. Packing only changes their temporary arrangement for efficient encoding, while their original image coordinates are retained for reassembly.

Following the same encoding and projection pipeline as full reassembly, each packed observation $\tilde I_b^l$ is processed as
{\footnotesize $\tilde T_b^l=\Pi_l(\phi(\tilde I_b^l))$}, for $b=1,\ldots,B$.
The encoded patch features are then restored to their original spatial locations as
{\footnotesize $L_{\mathcal S}=\mathcal{R}_{\mathcal S}\!\left(\{\tilde T_b^l\}_{b=1}^{B}\right)$},
where $\mathcal{R}_{\mathcal S}(\cdot)$ denotes selective reassembly.

Zero-filling unselected regions would discard useful scene context. We therefore fill them with the corresponding Global $G_0$ features:
\begin{equation}
\footnotesize
L(u)=
\begin{cases}
L_{\mathcal{S}}(u), & u\in\Omega_{\mathcal{S}},\\
G_0(u), & u\notin\Omega_{\mathcal{S}},
\end{cases}
\end{equation}
where $\Omega_{\mathcal S}$ denotes the spatial region covered by the selected patches. Selected regions thus contain high-fidelity local evidence from magnified observations, while unselected regions retain global context from $G_0$. CPSR therefore preserves a dense, globally aligned feature $L$ while reducing unnecessary high-resolution encoding under selective budgets.

Notably, CPSR estimates region importance directly in the image space rather than from encoded feature maps. Fine-grained cues of small objects may already be weakened by downsampling and feature compression, making feature-based selection more likely to overlook informative regions before high-resolution encoding. Operating directly on the image retains access to these local high-frequency cues and provides a low-cost signal for region selection before representation degradation.

\subsection{Density-Aware Adaptive Query Allocation (DA-AQA)}

In addition to spatially varying demands for high-resolution feature acquisition, aerial scenes also exhibit substantial variations in object density, leading to different decoder query requirements across images. Sparse images may require only a limited number of object hypotheses, whereas dense scenes can contain many targets. Using a fixed query budget can therefore lead to unnecessary decoding for sparse scenes or insufficient capacity for dense ones. To complement the spatial adaptation of Context-Preserved Selective Reassembly, we introduce \textbf{Density-Aware Adaptive Query Allocation (DA-AQA)}, which adapts the decoder query budget to image content using encoder proposal scores, without requiring an additional density predictor or auxiliary supervision.

Given the global-local feature set {\footnotesize $\mathcal{F}_{gl}=\{G_0,G_1,L\}$}, the encoder proposal stage produces candidate proposals with class logits. Let $\ell_{i,c}$ denote the class logit of proposal $i$ for category $c$. We compute the maximum class logit as {\footnotesize $z_i=\max_c \ell_{i,c}$}, rank proposals according to $z_i$, and retain the highest-scoring candidate pool of size $M$. For each retained proposal, we obtain a proposal score as {\footnotesize $s_i=\sigma(z_i)$}. Let $\{s_{(i)}\}_{i=1}^{M}$ denote the retained scores sorted in descending order. We compute the normalized cumulative proposal score mass as
{\footnotesize $C(k)=\sum_{i=1}^{k}s_{(i)}\big/\sum_{i=1}^{M}s_{(i)}$}.

DA-AQA first determines a raw image-specific query budget as
{\footnotesize $K_q^{\mathrm{raw}}=\min\{k\mid C(k)\ge\rho\}$},
\textit{i.e.}, the smallest number of retained proposals whose accumulated score reaches a predefined ratio $\rho$. When proposal confidence is concentrated in a small number of candidates, a smaller query budget is sufficient. In contrast, when high-confidence responses are distributed over more candidates, as often occurs in dense or ambiguous scenes, more proposals are required to cover the same fraction of proposal score mass. DA-AQA therefore adapts decoder computation to image content without explicit object-density prediction.

For stable and hardware-efficient computation, the final query budget is
{\footnotesize $K_q=\operatorname{clip}\!\left(\left\lceil K_q^{\mathrm{raw}}/r\right\rceil r,K_q^{\min},K_q^{\max}\right)$},
where $r$ denotes the prescribed rounding multiple. The top-$K_q$ proposal boxes initialize the decoder reference points, while the decoder content queries remain learnable embeddings. When samples with different $K_q$ are batched together, tensors are padded to the largest $K_q$ in the batch. The padded query slots may still pass through the decoder, but a validity mask excludes them from Hungarian matching and classification loss, while box and GIoU losses are computed only over valid matches. At inference, each image uses its own query budget. DA-AQA thus adapts the valid query count while preserving the original matching and losses.

\subsection{Training Objective and Implementation Details}

\paragraph{Training objective.}
LiG-DETR retains the original RF-DETR training objective and introduces no additional loss terms. The matched detection losses consist of classification, $\ell_1$ box regression, and GIoU losses. The local branch is used only for high-resolution feature acquisition: all ground-truth boxes and Hungarian matching remain in the original image coordinate system, without crop-level detection heads, matching, or additional local-branch supervision. CPSR and DA-AQA are integrated into the same end-to-end pipeline without extra supervision.

\paragraph{Implementation details.}
We use RF-DETR-L as the base detector with a trainable DINOv2 visual encoder initialized from pretrained weights. Full local reassembly uses a $2\times2$ partition. For CPSR, the input is divided into a $4\times4$ fine-grained patch grid ($N=4$), and selected patches are packed into $2\times2$ local observations ($G=2$) in descending order of their Laplacian scores. We evaluate $B=\{1,2,3,4\}$ packed local-observation budgets, corresponding to $K_p=\{4,8,12,16\}$ retained patches, respectively. 

LiG-DETR is trained using distributed data parallelism on 8 NVIDIA RTX 3090 GPUs for 72 epochs. We use AdamW with an initial learning rate of $1\times10^{-4}$ and a weight decay of $1\times10^{-4}$. The batch size is 1 per GPU with 16-step gradient accumulation, corresponding to an effective batch size of 128. Other optimization, data augmentation, and training settings follow RF-DETR.

\section{Experimental Results}

\subsection{Setup}

\paragraph{Comparison methods.}
We compare LiG-DETR with representative recent object detectors, including Dome-DETR~\citep{hu2025dome}, FBRT-YOLO~\citep{xiao2025fbrt}, SFS-DETR~\citep{jia2026sfs}, FSDC-DETR~\citep{liu2026fsdc}, YOLO-MS~\citep{chen2025yolo}, YOLOv13~\citep{lei2025yolov13}, YOLO26~\citep{sapkota2025yolo26}, LW-DETR~\citep{song2025lw}, D-FINE~\citep{peng2025d}, DEIMv2~\citep{huang2025real}, RT-DETRv4~\citep{liao2026rt}, and RF-DETR~\citep{robinson2026rf}. These methods cover diverse detector architectures and design paradigms, enabling a broad comparison of detection accuracy and computational efficiency.
Most non-RF baselines are evaluated at $640\times640$, following the default or commonly adopted resolution of their corresponding base detectors. For consistency, Dome-DETR and FSDC-DETR are also evaluated at $640\times640$, although their original VisDrone experiments use $800\times800$. Since Dome-DETR and FSDC-DETR are built upon D-FINE and DEIMv2, respectively, we adopt the same $640\times640$ setting as their corresponding base detectors. RF-DETR and LiG-DETR are evaluated at $704\times704$, retaining the native RF-DETR resolution for direct comparison with our base detector while keeping the input scale close to the $640\times640$ setting used by the other methods.

\begin{table}[h!] \vspace{-15pt}
\centering
\scriptsize
\setlength{\tabcolsep}{4.0pt}
\caption{Image distribution in DGDet-VisDrone.} 
\label{tab:dataset}
\begin{tabular}{l|c|c|c|c|c|c|c}
\hline
\textbf{Domain} & Cloudy Train/Test & Sunny & Dusk & Foggy & Night & Rainy & Total \\
\hline
\textbf{\# Images} & 2030~/~1016 & 2201 & 711 & 734 & 1892 & 45 & 8629 \\
\hline
\end{tabular}
\vspace{-15pt}
\end{table}

\paragraph{Datasets.}
We evaluate LiG-DETR on two aerial detection benchmarks and one cross-domain protocol: \textbf{VisDrone-DET2019}~\citep{du2019visdrone}, \textbf{AI-TOD-v2}~\citep{xu2022detecting}, and our proposed \textbf{DGDet-VisDrone}.
\textbf{VisDrone-DET2019} is a UAV-view detection benchmark containing 10,209 images, including 6,471 training images, 548 validation images, and 3,190 test images. It covers 10 categories, including pedestrians, vehicles, and bicycles, with scenes ranging from sparse to highly crowded environments. Following Dome-DETR, we train on the training split and report results on the validation split.
We further construct \textbf{DGDet-VisDrone}, a cross-domain protocol based on the annotated images of VisDrone-DET2019. The data are partitioned into six domains, as shown in Table~\ref{tab:dataset}. We use Cloudy as the source domain and split its images into source-training and source-test subsets at a ratio of 2:1. The remaining five domains are used exclusively as unseen target domains for cross-domain testing. 
\textbf{AI-TOD-v2} contains 28,036 images and 752,745 annotated instances, split into 11,214 training, 2,804 validation, and 14,018 test images. It contains 8 categories and is dominated by extremely small objects, with an average object size of 12.7 pixels and 86\% of instances smaller than 16 pixels. Following Dome-DETR, we report test-set results.

\paragraph{Evaluation protocol and metrics.}
We use COCO-style AP as the primary metric, averaged over IoU thresholds from 0.50 to 0.95 with a step of 0.05. Compared with AP$_{50}$ and AP$_{75}$, AP evaluates performance across different localization requirements and is particularly informative for small objects, where slight box errors can cause large IoU drops. For VisDrone-DET2019, we report AP$_s$, AP$_m$, and AP$_l$ for small, medium, and large objects, respectively. For AI-TOD-v2, we report AP$_s$ and AP$_m$. For efficiency, we report parameters, GFLOPs at the corresponding input resolutions, and FPS on a single RTX 3090 GPU with batch size 1.

\begin{table*}[t!]
\centering
\tiny
\setlength{\tabcolsep}{2.0pt}
\caption{Comparison on VisDrone-DET2019 val set (\%).}
\label{tab:VisDrone-DET2019}
\setlength{\tabcolsep}{2.0pt}
\renewcommand{\arraystretch}{1.08}
\resizebox{\textwidth}{!}{
\begin{tabular}{l|cccccccccc|>{\columncolor{gray!15}}cccc|ccc}
\hline
Methods & Pedestrain & People & Bicycle & Car & Van & Truck & Tricycle & Awning-tricycle & Bus & Motor & AP & AP$_s$ & AP$_m$ & AP$_l$ & Params. (M) & GFLOPs$\downarrow$ & FPS$\uparrow$\\
\hline
YOLOv13-L&25.9 &16.6 &11.1 &62.2 &37.2 &32.8 &22.0 &13.0 &52.4 &26.2 &29.9 &18.2 &41.4 &56.6 &27.6 &85.4 &37.9\\
YOLO-MS (YOLOv13-L)&25.4 &16.5 &10.2 &62.4 &37.5 &32.9 &22.1 &13.9 &50.4 &26.2 &29.8 &18.1 &41.3 &49.8 &31.2 &99.8 &42.9\\
YOLO26-L&28.4 &18.4 &10.2 &61.8 &35.7 &29.9 &21.3 &12.8 &48.4 &27.3 &29.4 &18.3 &38.9 &51.1 &24.8 &86.5 &58.3\\
YOLO-MS (YOLO26-L)&26.7 &17.8 &11.2 &60.2 &34.9 &29.4 &20.7 &11.4 &48.2 &27.1 &28.8 &20.3 &40.1 &48.6 &31.1 &141.6 &50.1\\
LW-DETR-M&20.0 &17.1 &9.9 &53.2 &30.7 &28.0 &19.4 &10.2 &40.6 &22.9&25.2 &14.7 &38.4 &64.8 &28.2 &83.8 &72.6\\
D-FINE-L&22.9 &18.9 &11.0 &58.3 &34.9 &27.3 &21.1 &10.8 &45.0 &26.9&27.7 &18.3 &39.3 &54.5 &31.0 &91.4 &32.8\\
DEIMv2-L&20.2 &16.0 &7.9 &57.0 &32.7 &25.7 &17.6 &11.1 &39.3 &23.8&25.1 &15.9 &36.0 &54.6 &32.0 &145.8  &35.9\\
RT-DETRv4-L&23.4 &19.1 &11.5 &59.7 &37.2 &29.7 &21.3 &12.1 &46.9 &28.2&28.9 &19.3 &40.4 &55.0 &31.0 &93.3 &42.5\\
Dome-DETR-L&34.7 &25.8 &19.3 &\textbf{65.6} &\textbf{43.9} &33.7 &28.1 &17.2 &\textbf{56.7} &\textbf{35.2}&30.4 &\textbf{48.5} &\textbf{61.8} &36.0 &36.0 &240.7 &14.8\\
FBRT-YOLO-L&20.5 &12.5 &7.9 &55.2 &27.7 &24.9 &17.5 &8.2 &42.3 &21.5 &28.9 &14.0 &36.4 &47.9 &14.6 &164.9 &104.5\\
SFS-DETR-S&29.8 &22.1 &13.7 &62.3 &35.1 &29.1 &22.8 &11.0 &49.5 &31.5 &30.7 &22.6 &41.5 &45.4 &28.3 &112.7 &40.4\\
FSDC-DETR-M&17.6 &14.2 &4.8 &54.9 &34.3 &22.6 &16.1 &9.0 &34.9 &21.2 &23.0 &16.8 &28.6 &37.2 &34.1 &121.5 &30.0\\
RF-DETR-L (baseline) &25.1 &20.7 &13.1 &57.2 &34.8 &31.3 &22.5 &14.3 &47.3 &27.7&29.4 &18.6 &43.4 &65.1 &33.9 &125.6 &51.6\\
\hline
LiG-DETR$_{B=1}$ &31.8&24.3&16.8&63.2&42.3&38.1&27.2&17.3&52.8&31.8&34.6&25.3&46.4&61.5 &38.7&284.0 &30.4\\
LiG-DETR$_{B=2}$ &33.7&25.7&18.7&63.9&43.3&39.3&28.9&17.9&54.1&33.3&35.9&26.5&48.0&60.9 &38.7&403.4 &24.9\\
LiG-DETR$_{B=3}$ &34.8&26.6&19.8&64.2&43.8&\textbf{40.9}&29.0&17.5&55.0&34.6&36.6&27.4&49.0&62.5 &38.7&522.9 &21.6\\
\textbf{LiG-DETR$_{B=4}$}  &\textbf{34.9}&\textbf{27.5}&\textbf{20.2}&64.4&43.6&40.0&\textbf{30.4}&\textbf{18.7}&56.2&35.1&\textbf{37.1}&27.6&49.9&\textbf{62.9}&38.7&642.4 &18.0\\
\hline
\end{tabular} 
}
\vspace{-15pt}
\end{table*}

\begin{table*}[t!]
\centering
\tiny
\setlength{\tabcolsep}{1.0pt}
\caption{Comparison with representative detectors on AI-TOD-v2 test set (\%).}
\begin{tabular}{l|cccccccc|>{\columncolor{gray!15}}ccc}
\hline
Methods & Airplane & Bridge & Storage-tank & Ship & Swimming-pool & Vehicle & Person & Wind-mill & AP & AP$_s$ & AP$_m$  \\
\hline
YOLOv13-L &\textbf{52.0} &20.4 &45.0 &47.6 &23.1 &36.0 &17.4 &9.0 &31.3 &26.2 &47.9   \\
YOLO-MS (YOLOv13-L) &48.3 &19.5 &43.9 &41.7 &22.1 &33.9 &15.8 &6.9 &29.0 &24.6 &46.7   \\
YOLO26-L  &47.4 &17.7 &45.7 &47.1 &23.9 &35.8 &18.0 &7.8 &30.4 &25.9 &45.0   \\
YOLO-MS (YOLOv26-L)  &45.1 &17.4 &43.7 &48.4 &\textbf{26.1} &34.0 &18.2 &7.5 &30.1 &22.8 &44.1 \\
LW-DETR-M &35.2 &10.7 &\textbf{56.6} &32.2 &20.3 &18.8 &12.7 &6.8 &20.4 &19.0 &42.2   \\
D-FINE-L  &38.8 &19.7 &34.1 &39.0 &19.5 &23.3 &14.6 &5.6 &24.3 &23.4 &42.6  \\
DEIMv2-L  &9.4&0.6&24.5&22.1&3.6&15.5&6.9&1.0 &10.4&10.1&18.5  \\
RT-DETRv4-L &37.4&15.1&32.8&32.7&19.2&22.0&12.1&4.8 &22.0&21.0&38.0  \\
Dome-DETR-L  &46.3 &\textbf{23.4} &44.3 &49.3 &23.5 &\textbf{37.1} &20.2 &7.4 &31.4 &\textbf{37.6} &45.4  \\
FBRT-YOLO-L  &48.8 &20.4 &43.9 &46.6 &22.4 &36.2 &17.3 &7.7 &30.4 &25.5 &45.8  \\
SFS-DETR-S&24.7 &16.6 &33.0 &32.3 &19.6 &22.9 &11.7 &3.6 &20.6 &19.7 &33.1 \\
FSDC-DETR-M&36.2 &12.0 &31.6 &32.6 &17.4 &21.7 &12.4 &5.0 &21.1 &20.1 &35.2 \\
RF-DETR-L (baseline) &38.2&13.6&33.8&38.6&22.7&26.0&16.5&9.6 &24.9&23.6&45.4  \\
\hline
\textbf{LiG-DETR} &46.9&21.9&43.7&\textbf{49.4}&25.2&34.5&\textbf{22.3}&\textbf{10.1}&\textbf{31.8}&30.8&\textbf{49.8}  \\
\hline
\end{tabular}\vspace{-15pt}
\label{tab:AITODv2}
\end{table*}

\subsection{Main Results}

\paragraph{Results on VisDrone-DET2019.}
Table~\ref{tab:VisDrone-DET2019} compares LiG-DETR with representative detectors on VisDrone-DET2019. LiG-DETR consistently improves over the RF-DETR-L baseline under different local high-resolution encoding budgets. Here, $B$ denotes the number of packed local observations in CPSR and reflects the local high-resolution encoding budget. With only one packed local observation ($B=1$), LiG-DETR achieves 34.6\% AP, improving RF-DETR-L by 5.2 points while running at 30.4 FPS. Increasing $B$ further improves detection accuracy, and LiG-DETR$_{B=4}$ reaches 37.1\% AP, outperforming RF-DETR-L by 7.7 points and achieving the highest AP among the compared methods.
The improvements are particularly pronounced for small and medium objects. Compared with RF-DETR-L, LiG-DETR$_{B=4}$ improves AP$_s$ from 18.6\% to 27.6\% and AP$_m$ from 43.4\% to 49.9\%, demonstrating the benefit of high-resolution local evidence for objects sensitive to spatial compression. Meanwhile, it retains strong large-object performance with 62.9\% AP$_l$. These results show that high-resolution local evidence substantially strengthens fine-grained object representations while retaining strong performance across object scales.

LiG-DETR also provides a controllable accuracy--efficiency trade-off through $B$. As $B$ increases from 1 to 4, AP improves from 34.6\% to 37.1\%, while FPS decreases from 30.4 to 18.0 due to additional local encoding. Even at $B=4$, LiG-DETR remains faster than Dome-DETR-L (18.0 vs.\ 14.8 FPS) while achieving 6.7 points higher AP. Unless otherwise specified, LiG-DETR$_{B=4}$ is used as the default configuration in subsequent experiments.

\begin{table*}[t!]
\centering
\scriptsize
\setlength{\tabcolsep}{2.0pt}
\caption{Generalization results on DGDet-VisDrone (\%).}
\resizebox{\textwidth}{!}{
\begin{tabular}{l|
>{\columncolor{gray!15}}c c c c|
>{\columncolor{gray!15}}c c c c|
>{\columncolor{gray!15}}c c c c|
>{\columncolor{gray!15}}c c c c|
>{\columncolor{gray!15}}c c c c|
>{\columncolor{gray!15}}c c c c|
>{\columncolor{gray!15}}c c c c}
\hline
\multirow{2}{*}{Methods}
& \multicolumn{4}{c|}{Cloudy (Source)}
& \multicolumn{4}{c|}{Sunny}
& \multicolumn{4}{c|}{Dusk}
& \multicolumn{4}{c|}{Foggy}
& \multicolumn{4}{c|}{Night}
& \multicolumn{4}{c|}{Rainy} 
& \multicolumn{4}{c}{Average} 
\\
\cline{2-29}
& AP & AP$_s$ & AP$_m$ & AP$_l$
& AP & AP$_s$ & AP$_m$ & AP$_l$
& AP & AP$_s$ & AP$_m$ & AP$_l$
& AP & AP$_s$ & AP$_m$ & AP$_l$
& AP & AP$_s$ & AP$_m$ & AP$_l$
& AP & AP$_s$ & AP$_m$ & AP$_l$
& AP & AP$_s$ & AP$_m$ & AP$_l$ \\
\hline
YOLOv13-L   &31.9 &17.5 &39.3 &49.3 &18.4 &8.3 &23.9 &30.8 &22.9 &10.9 &32.0 &41.2 &20.0 &9.0 &28.6 &41.4 &13.5 &4.6 &15.0 &23.0 &22.4 &9.1 &23.8 &60.1  &21.5 &9.9 &27.1 &41.0\\
YOLO-MS (YOLOv13-L) &30.7 &16.2 &39.0 &49.9 &18.7 &8.6 &24.7 &33.3 &22.8 &10.7 &32.5 &40.6 &20.1 &8.7 &29.8 &45.2 &12.4 &4.5 &15.1 &20.8 &22.5 &8.8 &22.9 &69.1  &21.2 &9.6 &27.3 &43.2\\
YOLOv26-L   &33.0 &18.7 &40.5 &50.1 &20.6 &9.7 &26.6 &36.6 &24.4 &11.9 &33.8 &48.1 &21.5 &10.0 &30.3 &47.4 &13.5 &4.6 &15.8 &25.9 &24.6 &10.3 &26.9 &64.2  &22.9 &10.9 &29.0 &45.4\\
YOLO-MS (YOLO26-L) &29.4 &18.2 &38.0 &46.9 &16.8 &9.2 &23.2 &31.3 &20.6 &11.0 &30.6 &36.3 &18.3 &10.1 &27.7 &42.7 &9.1 &4.1 &12.6 &19.1 &20.0 &9.8 &22.5 &62.0  &19.0 &10.4 &25.8 &39.7\\
LW-DETR-M   &26.2 &13.3 &36.4 &54.4 &19.3 &8.5 &28.4 &42.9 &21.6 &10.1 &33.2 &59.1 &18.7 &7.3 &30.7 &47.8 &16.5 &5.9 &22.2 &42.7 &24.6 &8.9 &25.7 &75.1  &21.2 &9.0 &29.4 &53.7\\
D-FINE-L    &25.6 &14.6 &34.7 &43.8 &17.4 &8.1 &24.4 &35.7 &20.2 &10.8 &30.4 &46.2 &17.1 &8.2 &26.6 &45.3 &13.6 &4.9 &18.0 &31.8 &22.4 &7.5 &23.4 &71.9  &19.4 &9.0 &26.3 &45.8\\
DEIMv2-L    &23.2 &12.5 &31.3 &41.3 &17.1 &8.2 &23.1 &33.4 &18.7 &10.0 &27.2 &31.7 &15.7 &7.2 &22.9 &37.2 &13.9 &5.1 &18.2 &32.1 &26.5 &30.7 &17.1 &20.9  &19.2 &12.3 &23.3 &32.8\\
RT-DETRv4-L  &24.0 &13.6 &32.4 &40.0 &16.9 &8.2 &23.0 &32.8 &18.7 &10.1 &27.9 &31.1 &16.4 &8.0 &24.6 &38.1 &11.8 &4.4 &15.7 &23.7 &26.1 &31.1 &18.4 &20.8  &19.0 &12.6 &23.7 &31.1\\
Dome-DETR-L &34.2 &\textbf{23.8} &42.1 &50.9 &23.9 &14.5 &29.7 &36.4 &26.6 &16.7 &35.0 &40.0 &24.9 &\textbf{15.7} &31.7 &43.8 &16.7 &7.3 &20.7 &31.1 &\textbf{32.9} &\textbf{35.3} &23.9 &23.0 &26.5 &\textbf{18.9} &30.5 &37.5\\
FBRT-YOLO-L &30.9 &14.3 &34.1 &42.7 &17.6 &6.1 &19.3 &27.1 &21.5 &7.5 &26.9 &34.8 &19.9 &6.2 &23.8 &42.6 &12.3 &2.9 &10.2 &19.0 &21.7 &6.7 &17.8 &58.0 &20.7 &7.3 &22.0 &37.4\\
SFS-DETR-S&29.6 &19.9 &38.0 &43.4 &19.6 &11.5 &26.6 &30.9 &22.9 &13.9 &32.8 &38.1 &20.6 &11.6 &31.6 &44.6 &10.7 &4.60 &14.5 &21.6 &21.3 &9.9 &23.7 &58.1 &20.7 &12.3 &28.7 &35.7 \\
FSDC-DETR-M &23.9 &13.9 &32.2 &38.5 &16.7 &9.0 &21.6 &28.7 &18.8 &10.4 &27.2 &30.4 &16.5 &8.6 &24.3 &36.3 &12.2 &6.2 &14.9 &21.9 &25.8 &27.7 &14.3 &18.8 &17.6 &9.6 &24.0 &31.2 \\
RF-DETR-L (baseline)   &29.5 &16.9 &39.9 &57.1 &22.5 &11.2 &32.2 &46.5 &25.1 &13.3 &37.2 &\textbf{60.6} &21.6 &9.5 &34.8 &56.5 &19.5 &7.5 &26.2 &43.7 &26.5 &9.5 &29.3 &\textbf{76.8}  &24.1 &11.3 &33.3 &56.9\\
\hline
\textbf{LiG-DETR}  &\textbf{34.8}&23.6&\textbf{45.8}&\textbf{59.4}&\textbf{28.4}&\textbf{16.9}&\textbf{39.4}&\textbf{49.3} &\textbf{30.2}&\textbf{18.9}&\textbf{43.0}&60.0 &\textbf{26.7}&15.3&\textbf{41.0}&\textbf{52.1} &\textbf{22.0}&\textbf{9.7}&\textbf{29.3}&\textbf{45.7} &32.4&14.6&\textbf{36.6}&75.8 &\textbf{29.1}&16.5&\textbf{39.2}&\textbf{57.1}\\
\hline
\end{tabular}
}
\vspace{-15pt}
\label{tab:DGDet-VisDrone}
\end{table*}

\paragraph{Results on AI-TOD-v2.}
Table~\ref{tab:AITODv2} reports the results on AI-TOD-v2. LiG-DETR achieves the highest overall AP of 31.8\%, outperforming RF-DETR-L by 6.9 points and surpassing all compared methods. This demonstrates the effectiveness of global-local reassembly for aerial imagery dominated by extremely small objects.
Compared with RF-DETR-L, LiG-DETR improves AP$_s$ from 23.6\% to 30.8\% and AP$_m$ from 45.4\% to 49.8\%, showing clear gains on both small and medium objects. Although Dome-DETR-L achieves higher AP$_s$, LiG-DETR obtains higher overall AP and AP$_m$, indicating a more balanced improvement across object scales. At the category level, LiG-DETR outperforms RF-DETR-L across all eight categories, further demonstrating the effectiveness of high-resolution local evidence for diverse aerial objects.

\paragraph{Results on DGDet-VisDrone.}
Small-object detection is particularly vulnerable to domain shifts because small instances provide limited texture, boundary, and shape cues. Changes in weather, illumination, and image quality can further perturb these already weak visual signals. We therefore evaluate LiG-DETR on DGDet-VisDrone to examine whether high-resolution local evidence improves cross-domain generalization to unseen domains.
As shown in Table~\ref{tab:DGDet-VisDrone}, LiG-DETR achieves the highest all-domain average AP of 29.1\%, outperforming RF-DETR-L by 5.0 points. More importantly, consistent improvements are observed across all five unseen target domains, including Sunny, Dusk, Foggy, Night, and Rainy, showing effectiveness across diverse appearance shifts. Compared with RF-DETR-L, LiG-DETR improves the all-domain average AP$_s$ from 11.3\% to 16.5\% and AP$_m$ from 33.3\% to 39.2\%. These results suggest that acquiring and reassembling high-resolution local evidence helps preserve more discriminative fine-grained information for small and medium objects under domain shifts.
Compared with Dome-DETR-L, LiG-DETR achieves higher average AP (29.1\% vs.\ 26.5\%), AP$_m$ (39.2\% vs.\ 30.5\%), and AP$_l$ (57.1\% vs.\ 37.5\%), while Dome-DETR-L obtains higher AP$_s$ (18.9\% vs.\ 16.5\%). Overall, LiG-DETR shows improved cross-domain generalization while retaining substantial gains on small and medium objects over RF-DETR-L.

\subsection{Ablation Study}

\setlength{\columnsep}{10pt}
\setlength{\intextsep}{4pt}
\begin{wraptable}{r}{0.53\columnwidth}\vspace{-14pt}
\centering
\tiny
\caption{Ablation study on VisDrone-DET2019 (\%).}
\label{tab:abla}
\resizebox{\linewidth}{!}{
\begin{tabular}{l|cccc}
\hline
Configuration & AP & AP$_s$ & AP$_m$ & AP$_l$ \\
\hline
RF-DETR-L ($\{G_1\}$)
&29.4&18.6&43.4&65.1\\
\hline
$\{G_1,L\}$
&32.9&23.3&45.3&59.0\\
\hline
$\{G_1,G_2\}$
&29.0&18.4&42.9&62.8\\
$\{G_0,G_2\}$
&30.0&20.1&43.0&62.5\\
$\{G_0,G_1\}$
&30.4&20.2&43.5&64.7\\
$\{G_0,G_1,G_2\}$
&30.1&20.2&43.0&62.9\\
\hline
$\{G_0,G_1\}+L$, fuse $G_0$
&35.0&25.3&48.0&63.0\\
$\{G_0,G_1\}+L$, fuse $G_1$
&33.5&24.2&45.8&58.7\\
$\{G_0,G_1\}+L$, fuse $\{G_0,G_1\}$
&35.5&26.6&46.8&61.8\\
$\{G_0,G_1,L\}$, no fusion
&35.5&26.4&47.8&62.9\\
\hline
+ DA-AQA (CPSR: Spatial Packing)
&37.0&27.8&49.3&63.2\\
+ DA-AQA (CPSR: Score-based Packing)
&37.1&27.6&49.9&62.9\\
\hline
\end{tabular}
}
\vspace{-8pt}
\end{wraptable}

\paragraph{Component ablation.}
Table~\ref{tab:abla} analyzes the contribution of each component in LiG-DETR. Starting from the RF-DETR-L baseline with only the encoder-native global feature $G_1$, adding the reassembled local feature improves AP from 29.4\% to 32.9\%, with AP$_s$ increasing from 18.6\% to 23.3\%. This shows that high-resolution local reassembly provides useful fine-grained evidence for small objects. However, AP$_l$ decreases from 65.1\% to 59.0\%, suggesting that local evidence combined with $G_1$ alone does not provide the best balance across object scales.

We next study different combinations of global feature levels without Local Reassembly. Here, $G_0$, $G_1$, and $G_2$ denote global features with strides 8, 16, and 32, respectively. Among the evaluated combinations, $\{G_0,G_1\}$ achieves the best overall performance, reaching 30.4\% AP and 20.2\% AP$_s$. This indicates that incorporating the higher-resolution $G_0$ is beneficial for small-object detection, while combining it with the encoder-native $G_1$ provides the best overall balance. Adding the coarser $G_2$ does not yield further gains, indicating limited benefit from very low-resolution features.

We then examine how the reassembled local feature $L$ should interact with the global features. Fusing $L$ into $G_1$ degrades performance, particularly AP$_l$, while fusion with $G_0$ or both $G_0$ and $G_1$ does not outperform treating $L$ as an independent feature level. Without explicit fusion, the model achieves 35.5\% AP, 47.8\% AP$_m$, and 62.9\% AP$_l$, supporting our choice to preserve $L$ as a separate feature level for joint decoding.

Finally, we examine the adaptive designs in LiG-DETR. With the global-local feature configuration fixed, introducing DA-AQA increases AP from 35.5\% to 37.1\%, demonstrating the benefit of content-adaptive query allocation. We further compare spatial-order and score-based packing in CPSR under DA-AQA. The two strategies perform nearly identically (37.0\% vs.\ 37.1\% AP), indicating that packing order has little effect once the encoded patch features are restored to their original spatial locations. We therefore adopt score-based packing by default. Table~\ref{tab:VisDrone-DET2019} evaluates the CPSR local-observation budget $B$ and its accuracy--efficiency trade-off.

\setlength{\columnsep}{10pt}
\setlength{\intextsep}{4pt}
\begin{wraptable}{r}{0.52\columnwidth}\vspace{-14pt}
\centering
\tiny
\setlength{\tabcolsep}{5pt}
\caption{High-resolution evidence acquisition (\%).}
\label{tab:highres_compare}
\resizebox{\linewidth}{!}{
\begin{tabular}{l|c|c|c|c|c}
\hline
\textbf{Method} & AP & AP$_s$ & Params. (M) & GFLOPs$\downarrow$ & FPS$\uparrow$ \\
\hline
RF-DETR-L & 29.4 & 18.6 & 33.9 & 125.6 & 51.6 \\
RF-DETR-L (2$\times$ Input) & 37.4  & 27.8  & 36.2 & 851.8 & 11.8 \\
LiG-DETR$_{B=1}$ & 34.6 & 25.3 & 38.7 & 284.0 & 30.4 \\
LiG-DETR$_{B=4}$ & 37.1 & 27.6 & 38.7 & 642.4 & 18.0 \\
\hline
\end{tabular}
}
\end{wraptable}
\paragraph{Comparison with full-image upscaling.}
We further compare LiG-DETR with directly doubling both spatial dimensions of the full-image input. As shown in Table~\ref{tab:highres_compare}, RF-DETR-L (2$\times$ Input) reaches 37.4\% AP, but uniformly increasing the full-image resolution raises the computation to 851.8 GFLOPs and reduces the speed to 11.8 FPS. Its parameter count also increases slightly under the higher-resolution input setting. In contrast, LiG-DETR$_{B=4}$ achieves comparable accuracy with 37.1\% AP while requiring only 642.4 GFLOPs and running at 18.0 FPS. LiG-DETR$_{B=1}$ further reduces computation to 284.0 GFLOPs while still achieving 34.6\% AP. These results show that local high-resolution evidence acquisition provides a more favorable accuracy--efficiency trade-off than full-image upscaling.

\setlength{\columnsep}{10pt}
\setlength{\intextsep}{4pt}
\begin{wraptable}{r}{0.55\columnwidth}\vspace{-14pt}
\centering
\tiny
\caption{Patch granularity ablation (\%).}
\label{tab:patch_granularity}
\vspace{-0pt}
\resizebox{\linewidth}{!}{
\begin{tabular}{l|cccc|ccc}
\hline
\textbf{\# Patches} & AP & AP$_s$ & AP$_m$ & AP$_l$
& Params. (M) & GFLOPs$\downarrow$ & FPS$\uparrow$ \\
\hline
4
&36.9&27.1&49.5&62.5
&38.7&642.4&19.0\\
16
&\textbf{37.1}&\textbf{27.6}&\textbf{49.9}&62.9
&38.7&642.4&18.0\\
64
&36.2&26.8&48.7&\textbf{63.9}
&38.7&642.4&12.4\\
\hline
\end{tabular}
}
\vspace{-2pt}
\end{wraptable}

\paragraph{Effect of patch granularity.}
Table~\ref{tab:patch_granularity} studies the effect of fine-grained patch granularity in CPSR. Here, \# Patches denotes the total number of fine-grained patches $N^2$ obtained by partitioning the input image. To isolate the effect of patch granularity, we fix the packed local-observation budget to $B=4$ and retain all patches. Accordingly, the packing grid size $G$ is set to 1, 2, and 4 for 4, 16, and 64 patches, respectively, so that all patches are packed into the same four local observations. This keeps the amount of local high-resolution encoding nearly unchanged across different granularities, with identical reported GFLOPs. Finer partitioning, however, introduces additional patch processing and reassembly overhead, leading to lower practical FPS.
Using 16 patches achieves the best overall performance with 37.1\% AP. Compared with 4 patches, increasing the granularity improves AP$_s$ and AP$_m$, suggesting that a moderate partition provides a better balance between localized fine-grained evidence and contextual continuity during packing and reassembly. Further increasing the number of patches to 64 reduces AP$_s$ and AP$_m$, possibly because overly fine partitioning fragments local context and introduces more patch boundaries. Although 64 patches achieves the highest AP$_l$, the 16-patch setting provides the best overall accuracy and is adopted by default.
\textbf{\textit{See Appendix~\ref{sec:visualization} for qualitative analysis and Appendix~\ref{sec:Future} for limitations and future work.}}

\section{Conclusion}
In this paper, we propose \textbf{LiG-DETR}, an Efficient Global-Local Reassembly framework for aerial object detection. LiG-DETR reformulates image slicing as high-resolution local feature acquisition and reassembles local evidence into the global coordinate space for unified end-to-end decoding, avoiding independent slice-level detection. Context-Preserved Selective Reassembly adaptively focuses high-resolution encoding on informative regions, while Density-Aware Adaptive Query Allocation dynamically adapts the decoder query budget. Extensive experiments show substantial improvements on small and medium objects while retaining strong large-object performance, together with a favorable accuracy-efficiency trade-off and improved cross-domain generalization.


\subsection*{AI use statement}
In this work, we used generative AI tools to assist with translation, language editing, and improving the readability and organization of the manuscript. We did not use generative AI tools for research design, method development, data processing, or interpretation of experimental results. All AI-assisted text was reviewed and revised by the authors. We take responsibility for the final content of this work, including any text produced with the aid of generative AI.

\subsection*{Ethics statement}
This work does not involve the collection of new human-subject data or personally identifiable information. All experiments are conducted on publicly available aerial object detection datasets or derived evaluation splits. The proposed DGDet-VisDrone protocol is constructed from the annotated images of VisDrone-DET2019, and the processed split files will be released for research purposes.

\subsection*{Reproducibility statement}
We provide detailed descriptions of the model architecture, training settings, evaluation protocols, and ablation configurations in the paper. The construction and domain splits of DGDet-VisDrone are also documented. The code and processed dataset splits will be made publicly available to facilitate reproducibility.

\bibliography{iclr2027_conference}
\bibliographystyle{iclr2027_conference}

\newpage

\appendix

\section*{Appendix}

\section{Qualitative Analysis}
\label{sec:visualization}

\begin{figure*}[h]
	\centering
	\includegraphics[width=1.0\linewidth]{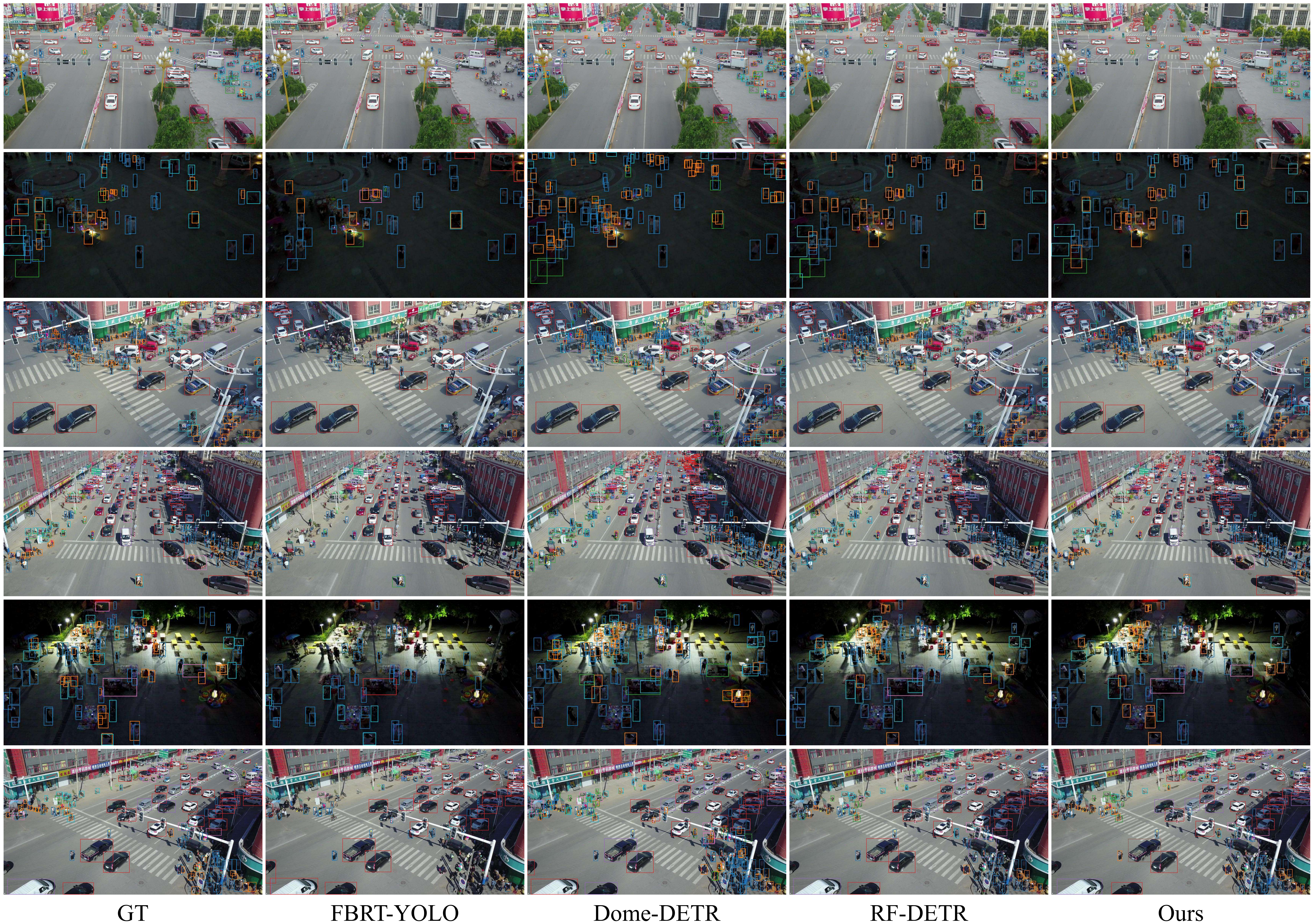}\vspace{-15pt}
	\caption{Qualitative comparison on VisDrone-DET2019. LiG-DETR produces more complete detections, particularly for small and densely distributed objects.}  %
	\label{fig:vision}
	\vspace{-0pt}
\end{figure*}

\paragraph{Qualitative results on VisDrone-DET2019.}
As shown in Fig.~\ref{fig:vision}, LiG-DETR produces more complete detections across diverse VisDrone scenes, including crowded intersections, dense pedestrian regions, and low-illumination environments. Compared with FBRT-YOLO, Dome-DETR, and RF-DETR, LiG-DETR recovers more small and distant instances that are easily missed when objects occupy only a few pixels or appear in densely clustered regions. The advantage is particularly evident in crowded traffic and nighttime scenes, where weak object responses and background interference make detection more difficult. In these cases, the predictions of LiG-DETR more closely follow the spatial distribution of the ground truth and exhibit fewer obvious missed detections. These qualitative results are consistent with the quantitative gains in AP$_s$ and AP$_m$, supporting our motivation that locally magnified high-resolution evidence can complement full-image representations with stronger fine-grained cues for challenging aerial objects.

\begin{figure*}[t]
	\centering
	\includegraphics[width=1.0\linewidth]{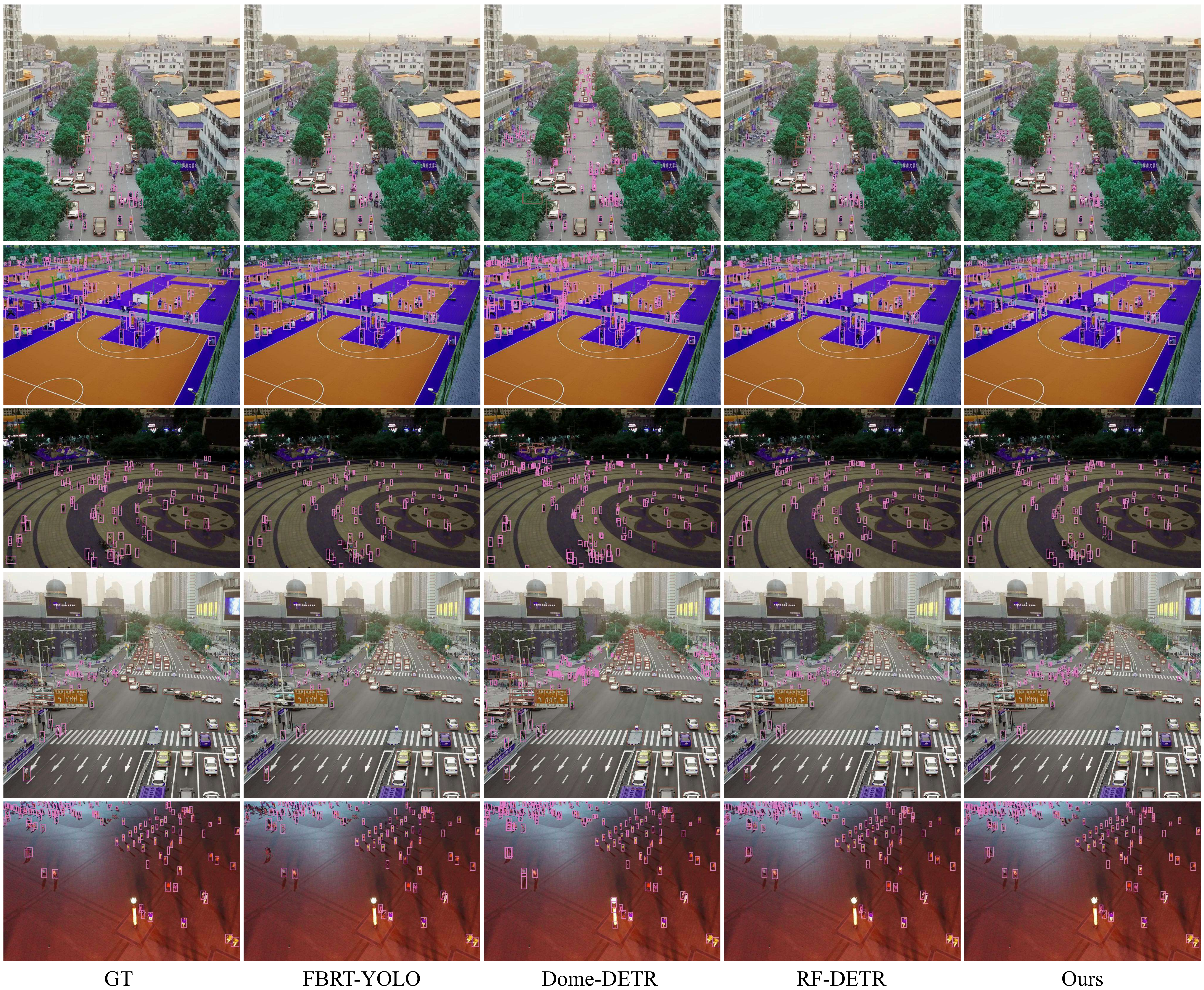}\vspace{-15pt}
	\caption{Qualitative comparison on AI-TOD-v2. LiG-DETR detects more extremely small objects and produces predictions closer to the ground-truth distribution.}  %
	\label{fig:vision2}
	\vspace{-0pt}
\end{figure*}

\paragraph{Qualitative results on AI-TOD-v2.}
As shown in Fig.~\ref{fig:vision2}, LiG-DETR also demonstrates stronger detection completeness on AI-TOD-v2, where objects are typically extremely small and sparsely or densely distributed over large aerial scenes. Compared with FBRT-YOLO, Dome-DETR, and RF-DETR, LiG-DETR recovers more weak and isolated instances as well as densely distributed tiny objects, particularly in regions where targets occupy only a very limited number of pixels. The compared detectors tend to miss more objects in such regions, whereas LiG-DETR produces predictions that better match the overall spatial distribution of the ground truth. The improvement is visible across different scene layouts, suggesting that the benefit of local high-resolution evidence is not limited to a particular background or density pattern. These observations are consistent with the improvements in AP$_s$ and AP$_m$, further indicating that acquiring fine-grained evidence before feature degradation helps preserve discriminative cues for extremely small aerial objects.

\section{Limitations and Future Work}
\label{sec:Future}

To the best of our knowledge, LiG-DETR is the first framework to reinterpret image slicing as high-resolution local feature acquisition and integrate the reassembled local evidence into a unified end-to-end detector without independent slice-level prediction. Nevertheless, this design still introduces additional computation for local high-resolution encoding. In particular, although increasing the local-observation budget $B$ consistently improves detection accuracy, the $B=4$ configuration remains noticeably more computationally expensive than the RF-DETR-L baseline. Further improving the efficiency of high-resolution local evidence acquisition therefore remains an important direction for future work.

Another limitation lies in the region-selection strategy of CPSR. The current design uses a lightweight Laplacian-based high-frequency score to identify informative regions before feature degradation. While effective and inexpensive, such low-level image cues are not always perfectly aligned with semantic object importance. Exploring more effective mechanisms for allocating high-resolution computation to truly informative regions is therefore another promising direction.



\end{document}